\documentclass[11pt,a4paper]{article}
\usepackage{times,latexsym}
\usepackage{url}
\usepackage[T1]{fontenc}

\usepackage[acceptedWithA]{tacl2018v2}
\usepackage{todonotes}
\usepackage{amsmath,amssymb}
\usepackage{graphicx}
\usepackage{tikz,graphics,color,fullpage,float,epsf}
\usepackage{enumitem}  
\usepackage{subcaption}

\usepackage{xspace,mfirstuc,tabulary}

\newif\iftaclinstructions
\taclinstructionsfalse 
\iftaclinstructions
\renewcommand{\confidential}{}
\renewcommand{\anonsubtext}{(No author info supplied here, for consistency with
TACL-submission anonymization requirements)}
\newcommand{\instr}
\fi

\iftaclpubformat 

\else

\fi

\newcommand*\samethanks[1][\value{footnote}]{\footnotemark[#1]}

\def\enxx{\textit{En}$\rightarrow$\textit{Any}~}
\def\xxen{\textit{Any}$\rightarrow$\textit{En}~}
\def\enxxt{\textit{En}$\rightarrow$\textit{Xx}~}
\def\yyent{\textit{Yy}$\rightarrow$\textit{En}~}
\def\yyxxt{\textit{Yy}$\rightarrow$\textit{Xx}~}
\def\beru{\textit{Be}$\rightarrow$\textit{Ru}~}
\def\yide{\textit{Yi}$\rightarrow$\textit{De}~}

\title{Massively Multilingual Neural Machine Translation \\in the Wild: Findings and Challenges}
\author{
\hspace{60px}Naveen Arivazhagan \thanks{~Equal contribution. Correspondence to \tt{navari,ankurbpn,orhanf@google.com}}\And
Ankur Bapna \samethanks[1]
 \And
\hspace{-85px}Orhan Firat \samethanks[1] \AND
Dmitry Lepikhin \And
Melvin Johnson\And
Maxim Krikun\And
Mia Xu Chen\And
Yuan Cao\AND
George Foster\And
Colin Cherry\And
Wolfgang Macherey\And
Zhifeng Chen\And
Yonghui Wu \AND
Google AI
}

\date{}

\begin{document}
\maketitle
\begin{abstract}

We introduce our efforts towards building a universal neural machine translation (NMT) system capable of translating between any language pair.
We set a milestone towards this goal by building a single massively multilingual NMT model handling 103 languages trained on over 25 billion examples. Our system demonstrates effective transfer learning ability, significantly improving translation quality of low-resource languages, while keeping high-resource language translation quality on-par with competitive bilingual baselines. We provide in-depth analysis of various aspects of model building that are crucial to achieving quality and practicality in universal NMT. While we prototype a high-quality universal translation system, our extensive empirical analysis exposes issues that need to be further addressed, and we suggest directions for future research.
\end{abstract}

\section{Introduction}

Sequence-to-sequence neural models (seq2seq)  \citep{kalchbrenner-blunsom:2013:EMNLP,sutskever2014sequence,D14-1179,bahdanau2014neural} have been widely adopted as the state-of-the-art approach for machine translation, both in the research community \citep{bojar2016findings,bojar2017findings,bojar-EtAl:2018:WMT1} and for large-scale production systems \citep{wu2016google,zhou2016deep,DBLP:journals/corr/CregoKKRYSABCDE16,hassan2018achieving}. As a highly expressive and abstract framework, seq2seq models can be trained to perform several tasks simultaneously \cite{luong2015multi}, as exemplified by multilingual NMT \citep{dong2015multi,firat2016multi,DBLP:journals/corr/HaNW16,johnson2017google} - using a single model to translate between multiple languages.

Multilingual NMT models are appealing for several reasons. Let's assume we are interested in mapping between $N$ languages; a naive approach that translates between any language pair from the given $N$ languages requires $O(N^2)$ individually trained models. When $N$ is large, the huge number of models become extremely difficult to train, deploy and maintain. By contrast, a multilingual model, if properly designed and trained, can handle all translation directions within a single model, dramatically reducing the training and serving cost and significantly simplifying deployment in production systems.

Apart from reducing operational costs, multilingual models improve performance on low and zero-resource language pairs due to joint training and consequent positive transfer from higher-resource languages \citep{zoph2016transfer,firat2016zero,nguyen-chiang:2017:I17-2,johnson2017google,neubig2018rapid,DBLP:journals/corr/abs-1903-00089, escolano2019towards,arivazhagan2019missing,hokamp2019evaluating}.
Unfortunately, this simultaneously results in performance degradation on high-resource languages due to interference and constrained capacity \citep{johnson2017google,tan2019multilingual}. Improving translation performance across the board on both high and low resource languages is an under-studied and challenging task.

While multilingual NMT has been widely studied, and the above benefits realized, most approaches are developed under constrained settings; their efficacy is yet to be demonstrated in real-world scenarios. In this work, we attempt to study multilingual neural machine translation \textit{in the wild}, using a massive open-domain dataset containing over 25 billion parallel sentences in 103 languages.

We first survey the relevant work in various areas: vocabulary composition, learning techniques, modeling and evaluation. In each area, we identify key determinants of performance and assess their impact in our setting. The result is a map of the landscape on this still largely unexplored frontier of natural language processing (NLP) and machine learning (ML). To the best of our knowledge, this is the largest multilingual NMT system to date, in terms of the amount of training data and number of languages considered at the same time. Based on experiments centered around different aspects of multilingual NMT we highlight key challenges and open problems on the way to building a real-world massively multilingual translation system.
\section{Towards Universal Machine Translation}
\label{sec:towards}

Enabling a single model to translate between an arbitrary language pair is the ultimate goal of universal MT. In order to reach this goal, the underlying machinery, \textit{the learner}, must model a massively multi-way input-output mapping task under strong constraints: 
a huge number of languages, different scripting systems, heavy data imbalance across languages and domains, and a practical limit on model capacity.
Now let us take a look at the problem from a machine learning perspective.

Machine learning algorithms are built on inductive biases in order to enable better generalization \cite{DBLP:books/daglib/0087929}. 
In the setting of multilingual NMT, the underlying inductive bias is that \textit{the learning signal from one language should benefit the quality of other languages} \cite{Caruana1997}.
Under this bias, the expectation is that as we increase the number of languages, the learner will generalize better due to the increased amount of information\footnote{Which can be sharing of semantic and/or syntactic structure, or ease of optimization via shared error signals etc.} added by each language (or task). This positive transfer is best observed for low-resource languages \cite{zoph2016transfer,firat2016multi,neubig2018rapid}. 
Unfortunately the above mentioned constraints prevent these gains from being progressively applicable: as we increase the number of languages with a fixed model capacity\footnote{Loosely measured in terms of the number of free parameters for neural networks.}, 
the positive/negative transfer boundary becomes salient, and high resource languages start to regress due to a reduction in per-task capacity.

From a function of mapping perspective, there are three common categories of multilingual NMT models in the literature, depending on the languages covered on the source and target sides: many-to-one, one-to-many and many-to-many models \cite{dong2015multi,firat2016multi,johnson2017google}. Many-to-one multilingual NMT models learn to map any of the languages in the source language set into the selected target language, usually chosen to be English due to the easy availability of parallel corpora with English on one side. Similarly, one-to-many multilingual NMT models aim to translate a single source language into multiple target languages. Many-to-one multilingual NMT can be categorized as a multi-domain\footnote{Note that we use the machine learning notion of domain here where each domain refers to a particular distribution of the input, while the target distribution remain unchanged.} learning problem \cite{Dredze:2010:MLC:1745449.1745451,joshi-etal-2012-multi,nam2015mdnet}, where the task of translating into a selected language remains the same, but the input distribution is different across source languages \cite{arivazhagan2019missing}. On the other hand, one-to-many multilingual NMT can be considered a multi-task problem \cite{Caruana1997,Thrun:1998:LL:296635,dong2015multi}, where each source-target pair is a separate task. Many-to-many translation is the super-set of both these tasks. Regardless of the number of languages considered on the source or the target side, \textbf{improvements in multilingual NMT are expected to arise from positive transfer between related domains and transferable tasks}.

Multi-domain and multi-task learning across a very large number of domains/tasks, with wide data imbalance, and very heterogeneous inter-task relationships arising from dataset noise and topic/style discrepancies,  
differing degrees of linguistic similarity, etc.,
make the path towards universal MT 
highly challenging.
These problems are typically approached individually and in constrained settings. Here we approach this challenge from the opposite end of the spectrum, determining what is possible with our current technology and understanding of NMT and ML, and probing for the effect of various design strategies in an extreme, real-world setting.

Our desired features of a truly multilingual translation model can be characterized as: 

\begin{itemize}    
    \item Maximum throughput in terms of the number of languages considered within a single model.
    \item Maximum inductive (positive) transfer towards low-resource languages.
    \item Minimum interference (negative transfer) for high-resource languages.
    \item Robust multilingual NMT models that perform well in realistic, open-domain settings.    
\end{itemize}

In the next sections we analyze different aspects of multilingual NMT, and investigate the implications of scaling up the dataset size and the number of languages considered within a single model. In each section we first discuss approaches described in recent literature, followed by our experimental setup, findings, and analysis. We also highlight some challenges and open problems identified in recent literature or as a result of our analyses. We start by describing our data setup, followed by an analysis of transfer and interference in the massively multilingual setting. We then analyze the pre-processing and vocabulary problems in multilingual NLP, discuss modeling approaches and the effect of increasing model capacity. We close with a discussion of evaluation and open problems in massively multilingual NMT.

\section{Data and Baselines}
\label{sec:data}
As with any machine learning problem, the quality and the amount of the data has significant impact on the systems that can be developed \cite{Goodfellow:2016:DL:3086952}. Multilingual NMT is typically studied using various public datasets or combinations of them. The most commonly used datasets include:
(\textit{i}) TED talks \cite{witted} which includes 59 languages \cite{Ye2018WordEmbeddings} with around 3k to 200k sentence pairs per language pair. (\textit{ii}) European parliamentary documents \cite{koehn2005epc} which include versions in 21 European languages having 1M to 2M sentence pairs. (\textit{iii}) The UNCorpus \cite{ZIEMSKI16.1195} is another multilingual dataset of parliamentary documents of United Nations, consisting of around 11 million sentences in 6 languages. (\textit{iv}) A compilation of the datasets used for the WMT News Translation shared task from 2005-19 \cite{tenyearsbojar,ws2018machinetranslation} covers a broader set of domains in around 15 languages, each containing between 10k to 50M sentence pairs. (\textit{v}) Other smaller parallel corpora for specific domains are indexed by OPUS \cite{tiedemann-2012-parallel} for various language pairs. (\textit{vi}) The Bible corpus is perhaps the most multilingual corpus publicly available, containing 30k sentences in over 900 languages \cite{tiedemann2018emerging}.

The results of various works on these datasets have greatly contributed to the progress of multilingual NMT. However, methods developed on these datasets and the consequent findings are not immediately applicable to real world settings outside those datasets due to their narrow domains, the number of languages covered or the amount of training data used for training. 

\begin{figure}[t!]
\begin{center}
\includegraphics[scale=0.2]{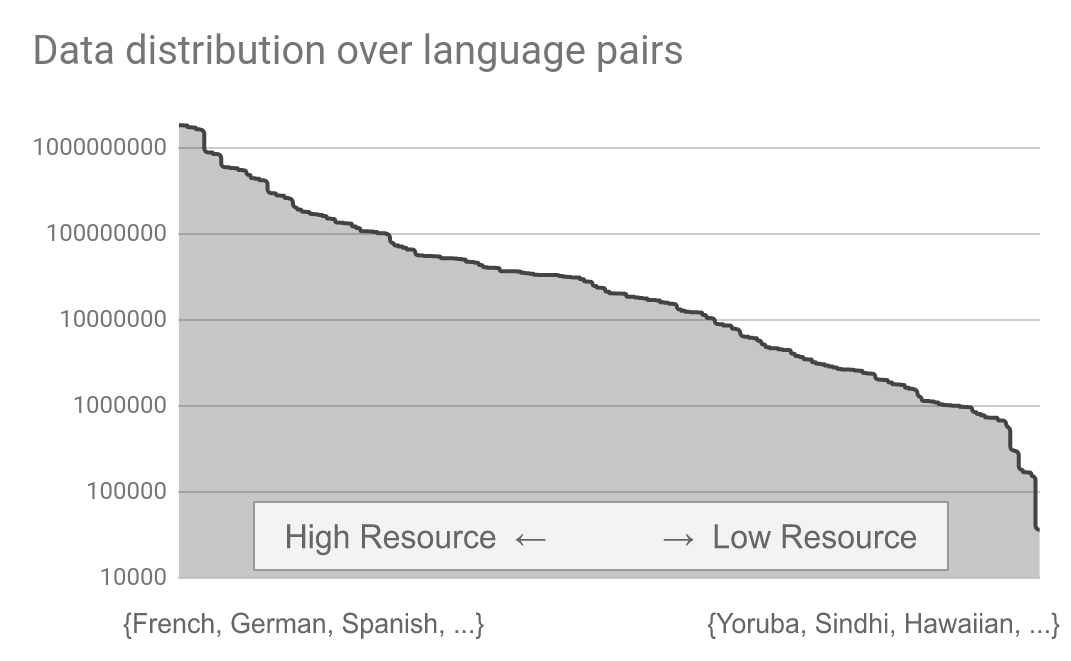}
\caption{Per language pair data distribution of the training dataset used for our multilingual experiments. The x-axis indicates the language pair index, and the y-axis depicts the number of training examples available per language pair on a logarithmic scale. Dataset sizes range from 35k for the lowest resource language pairs to 2 billion for the largest.}
\label{fig:data}
\end{center}
\end{figure}

\subsection{Our Data Setup}

Our problem significantly extends those studied by these previous works: we study multilingual NMT on a massive scale, using an in-house corpus generated by crawling and extracting parallel sentences from the web \cite{uszkoreit2010large}. This corpus contains parallel documents for 102 languages, to and from English, containing a total of 25 billion sentence pairs.\footnote{Limited to approximately this amount for experimentation.} \textbf{The number of parallel sentences per language in our corpus ranges from around tens of thousands to almost 2 billion}. Figure \ref{fig:data} illustrates the data distribution across language pairs for all 204 language pairs we study. The following specifics of our dataset distinguish our problem from previous work on multilingual NMT:

\begin{itemize}
\item Scale: Even on our lowest resource languages we often exceed the amount of data available in a majority of the previously studied datasets. Given the amount of data, techniques developed in relatively 
low-resource setups may not be as effective.
\item Distribution: The availability of quality parallel data follows a sharp power law, and data becomes increasingly scarce as we expand the scope of the system to more languages. There is a discrepancy of almost 5 orders of magnitude between our highest and our lowest resource languages. Balancing between these different language pairs now becomes a very challenging problem.
\item Domain and Noise: Having been mined from the web, our dataset spans a vast range of domains. However, such web crawled data is also extremely noisy; this problem gets worse in the multilingual setting where the level of noise might be different across different languages. While clean sources of parallel data are also available, they are often limited to narrow domains and high resource languages.
\end{itemize}

To summarize, the training data used in our study is drawn from 102 languages (+ English), exhibits a power-law in terms of number of training examples across language pairs, and spans a rich set of domains with varying noise levels---making our overall attempt as realistic as possible. Please see Table~\ref{tab:lang_ids} in the Appendix for the full list of languages.

\subsection{Experiments and Baselines}
Throughout this paper we perform several experiments on the training dataset described above, to highlight challenges associated with different aspects of multilingual models. We first train dedicated bilingual models on all language pairs to ground our multilingual analyses. We perform all our experiments with variants of the Transformer architecture \citep{vaswani2017attention}, using the open-source Lingvo framework \citep{lingvo}. For most bilingual experiments, we use a larger version of Transformer Big \citep{chen-EtAl:2018:Long1} containing around 375M parameters, and a shared source-target sentence-piece model (SPM) \cite{kudo-richardson-2018-sentencepiece} vocabulary with 32k tokens. We tune different values of regularization techniques (e.g. dropout \citep{srivastava2014dropout})  depending on the dataset size for each language pair. For most medium and low resource languages we also experiment with Transformer Base. All our models are trained with Adafactor \citep{shazeer2018adafactor} with momentum factorization, a learning rate schedule of (3.0, 40k),\footnote{(3.0, 40k) schedule is the shorthand for a learning rate of 3.0, with 40k warm-up steps for the schedule, which is decayed with the inverse square root of the number of training steps after warm-up.} and a per-parameter norm clipping threshold of 1.0. For Transformer Base models, we use a learning rate schedule of (2.0, 8k), unless otherwise necessary for low-resource experiments. 

In order to minimize confounding factors and control the evaluation set size and domain, we created our validation (development) and test sets as multi-way aligned datasets containing more than 3k and 5k sentence pairs respectively for all languages. For our bilingual baselines, BLEU scores are computed on the checkpoint with the best validation set performance, while we compute test BLEU on the final checkpoint (after training for around 1M steps) for our multilingual models, on true-cased output and references\footnote{We used an in-house implementation of mteval-v13a.pl from Moses to evaluate BLEU scores.}. For all our baselines we use a batch size of 1M tokens per-batch, by using large scale data parallelism over 16 TPUv3 chips \cite{46078}. We find that increasing the batch size offers noticeable improvements in model quality, while also significantly speeding up convergence.

\begin{figure}[t!]
\centering
  \begin{subfigure}[t]{\textwidth}
        \includegraphics[width=0.48\textwidth]{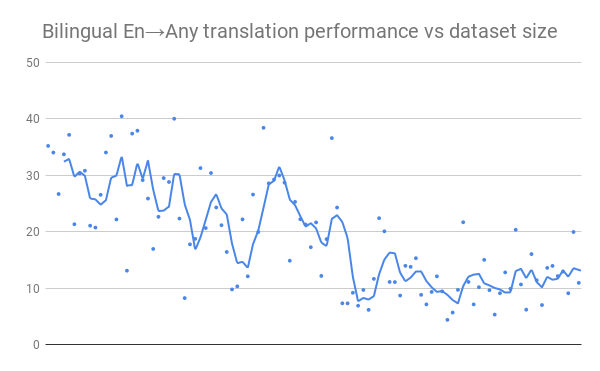}
    \end{subfigure}
    \begin{subfigure}[b]{\textwidth}
        \includegraphics[width=0.48\textwidth]{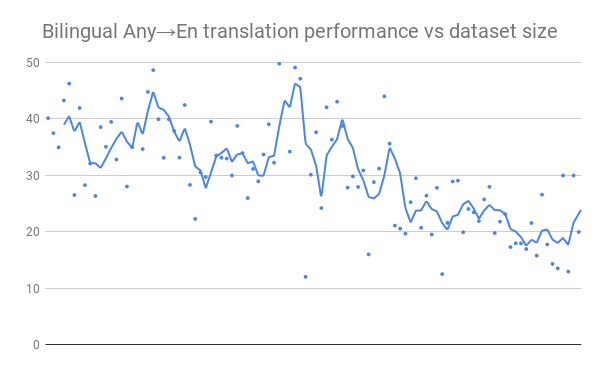}
    \end{subfigure}
\caption{Quality (measured by BLEU) of individual bilingual models on all 204 supervised language pairs, measured in terms of BLEU (y-axes). Languages are arranged in decreasing order of available training data from left to right on the x-axes (pair ids not shown for clarity). Top plot reports BLEU scores for translating from English to any of the other 102 languages. Bottom plot reports BLEU scores for translating from any of the other 102 languages to English. Performance on individual language pairs is reported using dots and a trailing average is used to show the trend.}
\label{fig:bilingual}
\end{figure}

\begin{table}[ht]
\centering
\begin{tabular}{l|c|c|c}
\textit{En}$\rightarrow$\textit{Any}     & High 25  & Med. 52  & Low 25 \\ \hline
Bilingual	&29.34	&17.50	&11.72 \\\hline \hline
\textit{Any}$\rightarrow$\textit{En}     & High 25  & Med. 52  & Low 25 \\ \hline
Bilingual	&37.61	&31.41	&21.63 \\\hline
\end{tabular}
\caption{Average translation quality (BLEU) of bilingual models over different groups of languages. High 25 refers to the top 25 languages by dataset size (left-most portion of Fig.~\ref{fig:data}), while low 25 refers to the bottom 25 (right-most portion of Fig.~\ref{fig:data}).}
\label{tab:bilingual}
\end{table}

We plot the BLEU scores for different language pairs in Figure~\ref{fig:bilingual}.
These results are also summarized in Table~\ref{tab:bilingual}. For brevity, we plot two main directions separately in different plots. When the source language is in English and we are translating \textit{from} English \textit{to} any other language, \textit{En}$\rightarrow$\textit{Any} is used for convenience, and similarly \textit{Any}$\rightarrow$\textit{En} for the opposite directions. We notice that translation performance on both \textit{En}$\rightarrow$\textit{Any} and \textit{Any}$\rightarrow$\textit{En} falls as the size of the training dataset decreases, as expected. In the next section we empirically analyze how multilingual models fare on the transfer-interference trade-off by using and comparing against the baselines introduced in this section.

\section{Learning}
\label{sec:learning}
Multilingual NMT is one of the largest multi-task problems being studied in academia or industry \cite{neubig2018rapid,DBLP:journals/corr/abs-1903-00089}, with hundreds of tasks (one per language pair) being learned in a single model. As is evident from Figure~\ref{fig:data}, multilingual NMT suffers from a severe data imbalance problem when studied in an unconstrained realistic setting.\footnote{The data-imbalance problem is also apparent in academical settings when multiple datasets are mixed, e.g. mixing TED talks with UN corpus.} While there is an abundance of data for some language pairs, making it difficult to go through even a single epoch over the entire dataset before model convergence, low resource languages suffer from data scarcity, making learning difficult. To make things worse, these learning problems might have varying levels of learning `difficulty' due to the linguistic properties of particular languages;
designing a learning algorithm to train a single model on all of these tasks simultaneously
is non-trivial.

In this section we will study the learning aspect of multilingual NMT, first examining the interaction between \textit{transfer} and \textit{interference} in our setup. Next, we will touch upon solutions to counter interference and lay out a set of future directions. And last, we will delve deeper into the transfer dynamics towards universal NMT.

\subsection{Transfer and Interference}
\label{subsec: transfer_and_interference}

Multitask learning \cite{Caruana1997} has been successfully applied to multiple domains, including NLP, speech processing, drug discovery and many others \citep{collobert08a,deng13new,ramsundar15massively,maurer16the,ruder17an,kaiser17one}. 
Other problems closely related to multitask learning include zero or few-shot learning \citep{Lake2011OneSL,romeraparedes15an,vinyals16matching,pan10survey}, meta-learning and life-long learning \citep{thrun95lifelong,silver13lifelong,chen2016lifelong,parisi18continual,golkan19continual,sodhani18on,lopez2017gradient}. 
Although these learning paradigms make different assumptions about the underlying problem setting, they share the common goal of leveraging the inductive bias and regularization from a set of tasks to benefit another set of tasks. 
This inductive transfer could be parallel or sequential.

Multilingual NMT has been studied under many of these settings to various extents. Most existing literature deals with sequential transfer, where the goal is to leverage a set of high resource tasks that are already mastered, to improve the performance on a new (predominantly) low resource task \cite{zoph2016transfer}. We consider the parallel learning problem, where \textbf{the goal is to learn a single multi-task model trained concurrently on all tasks and hence is capable of performing all tasks simultaneously} once the training is accomplished. 

In this section we investigate the effect of data imbalance across languages on these learning dynamics, particularly through the lens of \textit{transfer} and \textit{interference} \cite{Caruana1997,rosenstein2005transfer}. Reiterating from Section~\ref{sec:towards}, two desired characteristics of our universal machine translation model are (1) maximum (positive) transfer to low-resource languages and (2) minimum interference (negative transfer) for high-resource languages. Now let us examine the interaction between variables considered. For the baseline experiment, we start by following common conventions in literature on multilingual NMT \cite{FIRAT2017236,lee2017fully,johnson2017google}. We compare the performance of two training approaches against bilingual baselines following two strategies: 

\begin{enumerate}[label=(\roman*)]
\item all the available training data is combined as it is, with the data distribution in Figure~\ref{fig:data}, 
\item we over-sample (up-sample) low-resource languages so that they appear with equal probability in the combined dataset.
\end{enumerate}  

\begin{figure}[h!]
\centering
  \begin{subfigure}[t]{\textwidth}
        \includegraphics[width=0.48\textwidth]{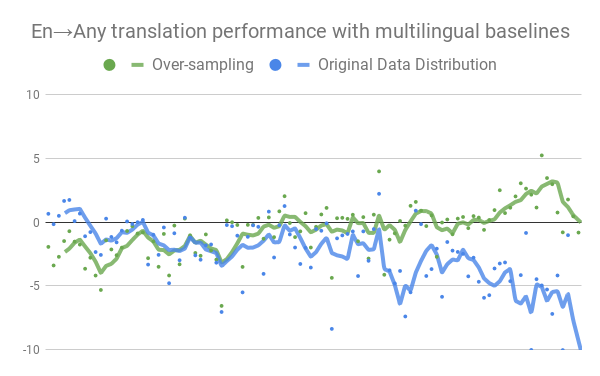}
    \end{subfigure}
    \begin{subfigure}[b]{\textwidth}
        \includegraphics[width=0.48\textwidth]{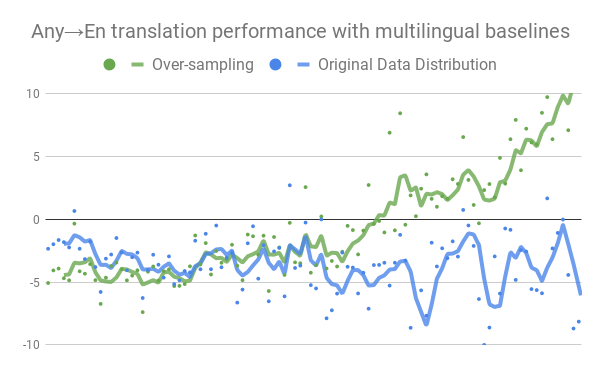}
    \end{subfigure}
\caption{Effect of sampling strategy on the performance of multilingual models. From left to right, languages are arranged in decreasing order of available training data. While the multilingual models are trained to translate both directions, \textit{Any}$\rightarrow$\textit{En} and \textit{En}$\rightarrow$\textit{Any}, performance for each of these directions is depicted in separate plots to highlight differences. Results are reported relative to those of the bilingual baselines (\ref{fig:bilingual}). Performance on individual language pairs is reported using dots and a trailing average is used to show the trend. The colors correspond to the following sampling strategies: (i) Blue: original data distribution, (ii) Green: equal sampling from all language pairs. Best viewed in color.}
\label{fig:baseline}
\end{figure}

In order to guide the translation with the intended target language, we pre-pend a target language token to every source sequence to be translated \citep{johnson2017google}. Further, to study the effect of \textit{transfer} and \textit{interference} at its limit, we shared a single encoder and decoder across all the language pairs. During training, mini-batches are formed by randomly sampling examples from the aggregated dataset following strategies (i) or (ii), as described above. We train a single Transformer-Big with a shared vocabulary of 64k tokens, all Transformer dropout options turned on with probability 0.1, and the same values as the bilingual baselines used for other hyper-parameters. We use batch sizes of 4M tokens for all our multilingual models to improve the rate of convergence. All Transformer-Big runs utilize data parallelism over 64 TPUv3 chips. The results are depicted in Figure~\ref{fig:baseline}.

The performance of these two models highlights the trade-off between \textit{transfer} and \textit{interference}. If we sample equally from all datasets by over-sampling low-resource languages (strategy (ii)), we maximize transfer (right-most portion of Figure~\ref{fig:baseline}) and beat our bilingual baselines by large margins, especially in the \textit{Any}$\rightarrow$\textit{En} direction. However, this also has the side-effect of significantly deteriorated performance on high resource languages (left-most portion of Figure~\ref{fig:baseline}). On the other hand, sampling based on the true data distribution (strategy (i)) retains more performance on high resource languages, at the cost of sacrificing performance on low resource languages. We also note that \textbf{the transfer-interference trade-off is more pronounced in the \textit{Any}$\rightarrow$\textit{En} direction: the cost-benefit trade-off between low and high-resource languages is more severe than that for the \textit{En}$\rightarrow$\textit{Any} direction.}

Another interesting finding is the performance deterioration on high resource languages when translating from \textit{Any}$\rightarrow$\textit{En}, contrary to existing results in multilingual NMT \cite{firat2016multi,johnson2017google}, exaggerated by the limited model capacity and the scale and imbalance of our dataset.
All these observations again underscore the difficulty in multitask learning, especially when hundreds of tasks are to be learned simultaneously, each of which may come from a different distribution. Although~\citep{maurer16the} demonstrates the benefit of multitask learning when invariant features can be shared across all tasks, such a premise is not guaranteed when hundreds of languages from different families are jointly considered.

\subsection{Countering Interference: Baselines and Open Problems}
\label{sec:counter}

The results in Figure~\ref{fig:baseline} indicate that, in a large multi-task setting,
high resource tasks are starved for capacity while low resource tasks benefit significantly from transfer, and the extent of interference and transfer are strongly related. However, this trade-off could be controlled by 
applying proper data sampling strategies.
 
\begin{figure}[h!]
\centering
\includegraphics[width=0.4\textwidth]{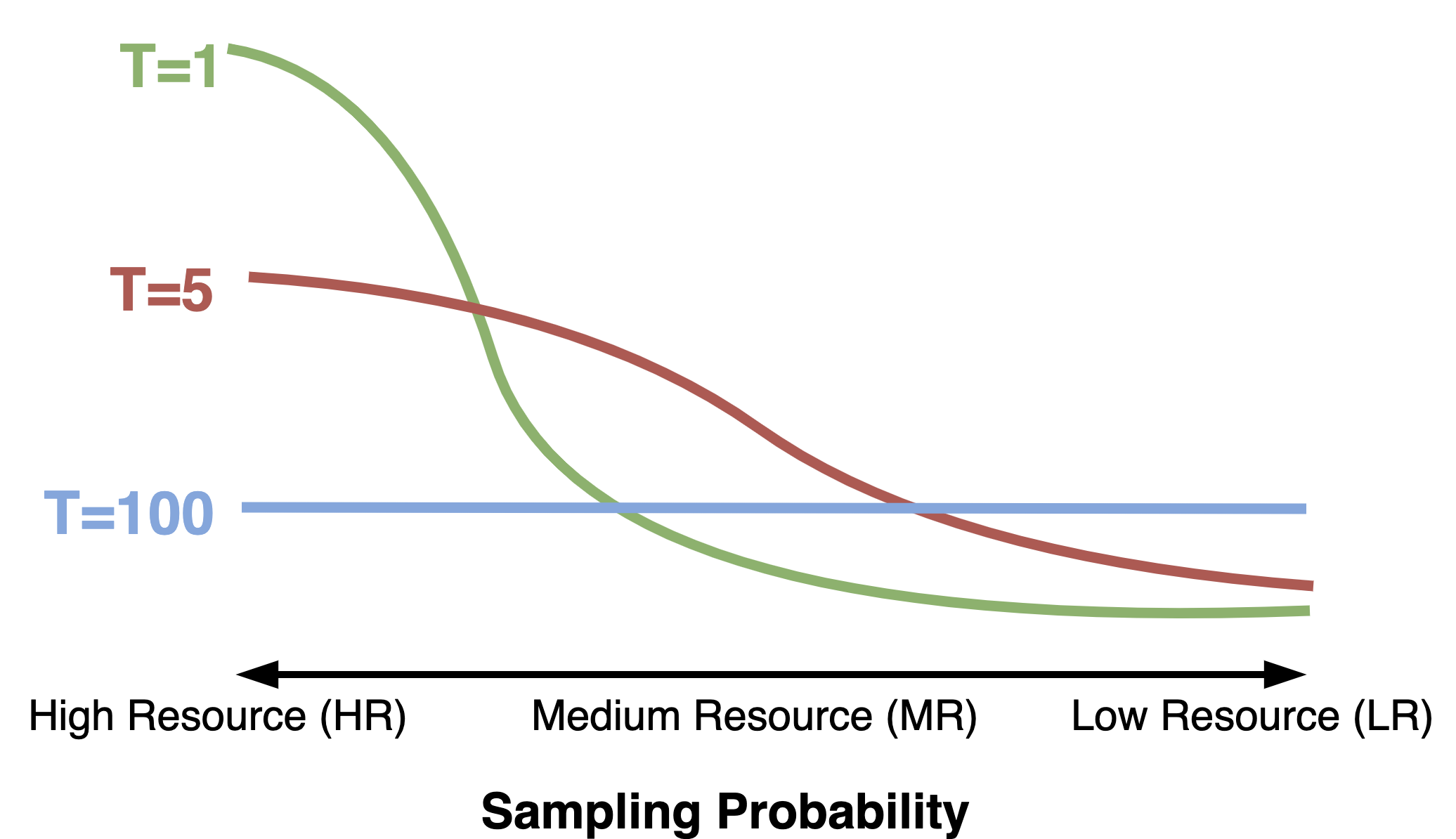}
\caption{Temperature based data sampling strategies overlaid on the data distribution.}
\label{fig:temperature_sampling}
\end{figure}

To enable more control over sampling, we investigate batch balancing strategies \cite{FIRAT2017236,lee2017fully}, along the lines of the temperature based variant used for training multilingual BERT \cite{devlin2018bert}. For a given language pair, $l$, let $D_l$ be the size of the available parallel corpus. Then if we adopt a naive strategy and sample from the union of the datasets, the probability of the sample being from language $l$ is $p_l=\frac{D_l}{\Sigma_kD_k}$. However, this strategy would starve low resource languages. To control the ratio of samples from different language pairs, we sample a fixed number of sentences from the training data, with the probability of a sentence belonging to language pair $l$ being proportional to $p_l^{\frac{1}{T}}$, where $T$ is the sampling temperature. As a result, $T=1$ corresponds to true data distribution and $T=100$ corresponds to (almost) equal number of samples for each language (close to a uniform distribution with over-sampled low-resource languages). Please see Figure~\ref{fig:temperature_sampling} for an illustration of the effect of temperature based sampling overlaid on our dataset distribution.

\begin{figure}[t!]
\centering
   \begin{subfigure}[t]{\textwidth}
   \includegraphics[width=0.48\textwidth]{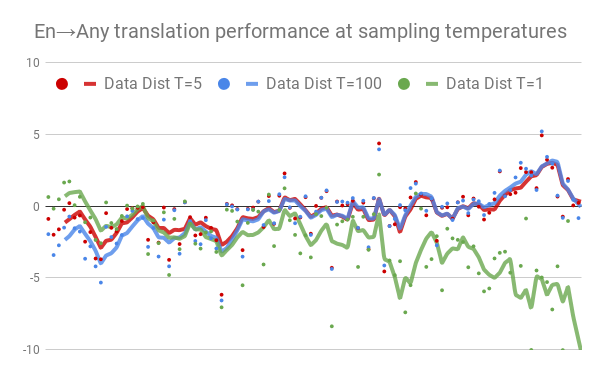}
\end{subfigure}
\begin{subfigure}[b]{\textwidth}
   \includegraphics[width=0.48\textwidth]{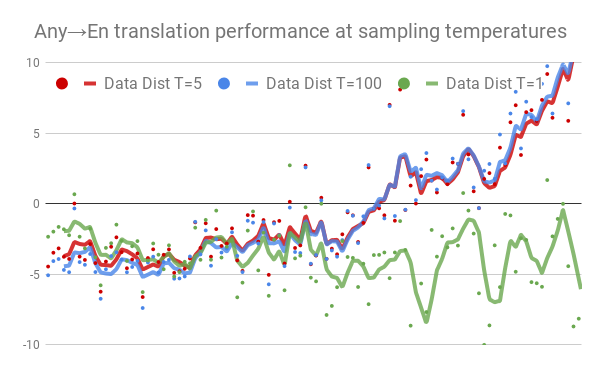}
\end{subfigure}
\caption{Effect of varying the sampling temperature on the performance of multilingual models. From left to right, languages are arranged in decreasing order of available training data. Results are reported relative to those of the bilingual baselines (\ref{fig:bilingual}). Performance on individual language pairs is reported using dots and a trailing average is used to show the trend. The colors correspond to the following sampling strategies: (i) Green: True data distribution ($T=1$) (ii) Blue: Equal sampling from all language pairs ($T=100$) (iii) Red: Intermediate distribution ($T=5$). Best viewed in color.}
\label{fig:temperature}
\end{figure}

We repeat the experiment in Section~\ref{subsec: transfer_and_interference} with temperature based sampling, setting $T=5$ for a balanced sampling strategy, and depict our results in Figure~\ref{fig:temperature}. Results over different language groups by resource size are also summarized in Table~\ref{tab:temperature}. We notice that the balanced sampling strategy improves performance on the high resource languages for both translation directions (compared to $T=100$), while also retaining high transfer performance on low resource languages. However, performance on high and medium resource languages still lags behind their bilingual baselines by significant margins.

We unveiled one of the factors responsible for interference while training massively multilingual NMT models under heavy dataset imbalance, and hinted that an appropriate data sampling strategy can potentially mitigate the interference. But the imbalance in dataset across tasks is not the only variable interacting with the \textit{transfer} - \textit{interference} dilemma. In all the experiments described above, multilingual models
have the same capacity as the baselines, a strategy which could be interpreted as reducing their per-task capacity.
To highlight the exacerbating effect of interference with increasing multilinguality, we train three additional models on a growing subset of 10, 25, and 50 languages. The specific languages are chosen to get a mixed representation of data size, script, morphological complexity and inter-language relatedness. Results for the 10 languages, with a data sampling strategy $T=5$, that are common across all subsets are reported in Figure \ref{fig:lang} and clearly highlight how \textbf{performance degrades for all language pairs, especially the high and medium resource ones, as the number of tasks grows}.

\begin{table}[t]
\centering
\begin{tabular}{l|c|c|c}
\textit{En}$\rightarrow$\textit{Any}      & High 25  & Med. 52  & Low 25 \\ \hline
Bilingual	&29.34	&17.50	&11.72 \\\hline 
T=1	& 28.63	&15.11	&6.24 \\\hline
T=100	&27.20	&16.84	&12.87 \\\hline
T=5	&28.03	&16.91	&12.75 \\\hline\hline
\textit{Any}$\rightarrow$\textit{En}      & High 25  & Med. 52  & Low 25 \\ \hline
Bilingual	&37.61	&31.41	&21.63 \\\hline
T=1	&34.60	&27.46	&18.14 \\\hline
T=100	&33.25	&30.13	&27.32 \\\hline
T=5	&33.85	&30.25	&26.96 \\\hline
\end{tabular}
\caption{Average translation quality (BLEU) of multilingual models using different sampling temperatures, over different groups of languages. High 25 refers to the top 25 languages by dataset size, while low 25 refers to the bottom 25.}
\label{tab:temperature}
\end{table}

While simple data balancing/sampling strategies might reduce the effects of interference without reducing transfer, our experiments also highlight a few research directions worth further exploration. Most notably, Should we be using the same learning algorithms for multilingual and single language pair models? Can we still rely on data-sampling heuristics when the number of tasks are excessively large? We highlight a few open problems along these lines:

\begin{figure}[h!]
\centering
   \begin{subfigure}[t]{\textwidth}
   \includegraphics[width=0.48\textwidth]{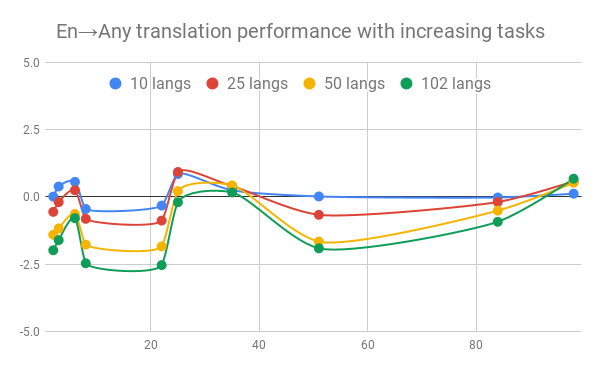}
\end{subfigure}
\begin{subfigure}[b]{\textwidth}
   \includegraphics[width=0.48\textwidth]{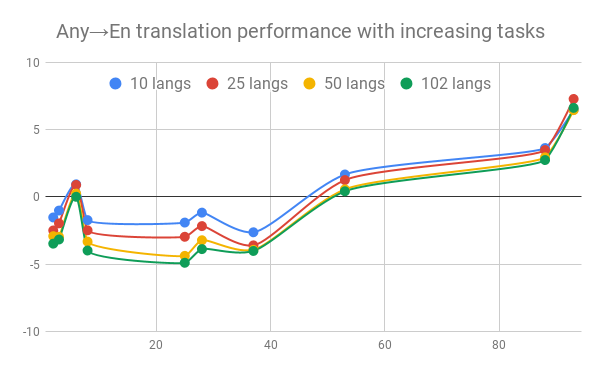}
\end{subfigure}
\caption{Effect of increasing the number of languages on the translation performance of multilingual models. From left to right, languages are arranged in decreasing order of available training data. Results are reported relative to those of the bilingual baselines (\ref{fig:bilingual}). The colors correspond to the following groupings of languages: (i) Blue: 10 languages $\leftrightarrow$ En, (ii) Red: 25 languages $\leftrightarrow$ En, (iii) Yellow: 50 languages $\leftrightarrow$ En (yellow), and (iv) Green: 102 languages $\leftrightarrow$ En. Note that, we only plot performance on the 10 languages common across all the compared models while keeping the x-axis intact for comparison with other plots. Best viewed in color.}
\label{fig:lang}
\end{figure}

\paragraph{Task Scheduling:} Scheduling tasks has been widely studied in the context of multitask learning and meta-learning, but remains relatively under-explored for multilingual NMT \citep{bengio2009curriculum,Pentina_2015_CVPR}. The scheduling of the tasks, or the scheduling of the corresponding data associated with the task can be studied under two distinct categories, static and dynamic (curriculum learning). Temperature based sampling or co-training with related languages to improve adaptation \citep{zoph2016transfer,neubig2018rapid} fall within the class of static strategies. On the other hand, dynamic or curriculum learning strategies refine the ratio of tasks  simultaneously with training, based on metrics derived from the current state of the learner \citep{graves2017automated}. In the context of NMT, \cite{DBLP:journals/corr/abs-1903-00041} learn a RL agent to schedule between different noise levels in the training data, while \cite{DBLP:journals/corr/abs-1903-09848} combined heuristics into a data curriculum. \cite{Kiperwasser17} designed schedules favoring one target language and \cite{seb2018clw} learn an adaptive scheduler for multilingual NMT. Similar approaches could be extended in the context of learning different language pairs in massively multilingual settings.

\paragraph{Optimization for Multitask Learning:} While task scheduling alters the data distribution or the dynamics of the data distribution seen by the \textit{learner}, optimization algorithms, regularization and loss formulations determine how these examples effect the learning process. While most literature in multilingual NMT, including our work, relies on the same monolithic optimization approach for single and multitask models,\footnote{One adaptive optimizer e.g. Adam, Adafactor with shared accumulators for gradient moments across all tasks.} this choice might be far from optimal. There is no dearth of literature exploring loss formulations or regularization techniques that unravel and exploit task relatedness, reformulate loss functions to account for adaptation and exploit meta-learning approaches in multitask models \citep{vilalta2002perspective,zhang2017survey}. Applying optimization approaches designed specifically for multitask models to multilingual NMT might be a fruitful avenue for future research.

\subsection{Understanding Transfer}
From our experiments on data sampling, we notice that multilingual training with shared weights helps promote transfer to low-resource languages. However, these improvements are imbalanced in how they affect languages when translating to or from English. To better understand \textit{transfer} in multilingual models we individually inspect three different settings: 1) translation from English (\textit{En}$\rightarrow$\textit{Any}), 2) translation to English (\textit{Any}$\rightarrow$\textit{En}), and 3) translation between non-English language pairs (\textit{Any}$\rightarrow$\textit{Any}). 

\begin{figure}[t!]
\centering
  \begin{subfigure}[t]{\textwidth}
  \includegraphics[width=0.48\textwidth]{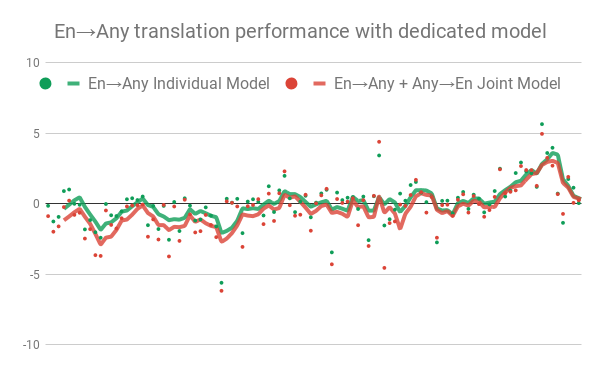}
\end{subfigure}
\begin{subfigure}[b]{\textwidth}
  \includegraphics[width=0.48\textwidth]{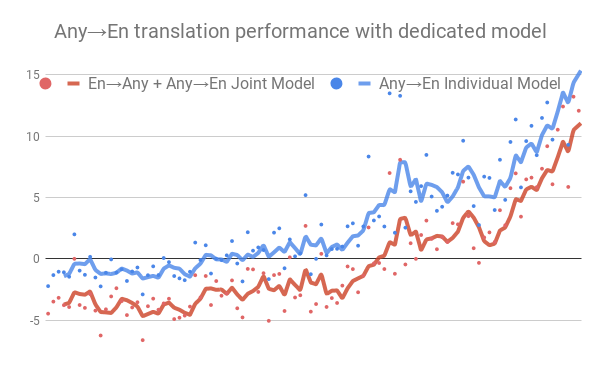}
\end{subfigure}
\caption{Results comparing the performance of models trained to translate English to and from all languages to two separate from and to English models. From left to right, languages are arranged in decreasing order of available training data. Results are reported relative to those of the bilingual baselines (\ref{fig:bilingual}). The colors correspond to the following models: (i) Green: dedicated (individual) \enxx model for top plot, (ii) Blue: dedicated \xxen model for bottom plot, and (iii) Red: shared model for both \xxen and \enxx. Best viewed in color.}
\label{fig:enxxen}
\end{figure}

We compare the performance of our model trained on all language pairs against two models: (i) An identical model trained on all \textit{En}$\rightarrow$\textit{Any} tasks, the one-to-many setup, and (ii) An identical model trained on all \textit{Any}$\rightarrow$\textit{En} tasks, the many-to-one setup. We depict the performance of these models in Figure~\ref{fig:enxxen} (and summarize in Table~\ref{tab:enxxen}). We notice that the many-to-one \textit{Any}$\rightarrow$\textit{En} model achieves huge improvements over our bilingual baselines for all low-resource languages (right-most portion of Figure~\ref{fig:enxxen}). On the other hand, for the one-to-many \textit{En}$\rightarrow$\textit{Any} model, we notice lesser deterioration in the performance on high resource languages, while the performance on low resource languages does not improve by much. 

\begin{table}[ht]
\centering
\begin{tabular}{l|c|c|c}
\enxx      & High 25  & Med. 52  & Low 25 \\ \hline
Bilingual	& 29.34	& 17.50	& 11.72 \\\hline 
\textit{All}$\rightarrow$\textit{All}	& 28.03	& 16.91	& 12.75 \\\hline
\enxx	& 28.75	& 17.32	& 12.98 \\\hline \hline
\xxen      & High 25  & Med. 52  & Low 25 \\ \hline
Bilingual	& 37.61	& 31.41	& 21.63 \\\hline
\textit{All}$\rightarrow$\textit{All}	& 33.85	& 30.25	& 26.96\\\hline
\xxen	& 36.61	& 33.66	& 30.56 \\\hline
\end{tabular}
\caption{Average translation quality (BLEU) of multilingual models trained on differing groups of languages. High 25 refers to the top 25 languages by dataset size, while low 25 refers to the bottom 25. \textit{All}$\rightarrow$\textit{All} reports the performance of the multilingual model trained on all language pairs, \enxx was trained on all language pairs with English as the source and \xxen was trained on all language pairs with English as the target.}
\label{tab:enxxen}
\end{table}

\begin{table*}[h!]
\centering
\begin{tabular}{l|c|c|c|c|c|c}
          & $De\rightarrow{}Fr$  & $Be\rightarrow{}Ru$  & $Yi\rightarrow{}De$  & $Fr\rightarrow{}Zh$  & $Hi\rightarrow{}Fi$ & $Ru\rightarrow{}Fi$ \\ \hline
10 langs  & 11.15 & 36.28 & 8.97 & 15.07 & 2.98 & 6.02 \\ \hline
102 langs & 14.24 & 50.26 & 20.00 & 11.83 & 8.76 & 9.06
\end{tabular}
\caption{Effect of increasing the number of languages on the zero-shot performance of multilingual models.}
\label{tab:zs}
\end{table*}

This discrepancy between the transfer for \xxen and \enxx can be better understood under the characterization of the many-to-one \xxen model as a multi-domain model, where each source language constitutes a separate domain, and the one-to-many \enxx model as a multi-task model, with each target language representing a separate task (Section~\ref{sec:towards}). This formulation of multilingual NMT helps explain the aforementioned observations, which suggest that multilingual models might be more amenable to transfer across input domains than transfer across tasks. Simple joint training does not do much to benefit one-to-many multilingual NMT; while some improvements may arise from the model seeing much more English source data, there is little to no transfer occurring at the task/decoder level. On the other hand, it is much easier for many-to-one multilingual models to reap the benefits of joint training when the inputs are from different domains: the output distribution remains the same, hence learning a much stronger English language model, without any interference from the other target languages or tasks. This is also reflected in other works on low-resource machine translation where \xxen typically benefits the most \citep{zoph2016transfer, nguyen-chiang:2017:I17-2,gu-EtAl:2018:N18-1,gu2018meta}.

Another strong indicator of transfer in multilingual models is the quality on zero-shot translation. Multilingual models possess a unique advantage over single task models, in that they are capable of translating between any pair of supported input and output languages, even when no direct parallel data is available \citep{firat2016zero}. However, without supervision from parallel data between non-English language pairs, zero-shot translation quality often trails the performance of pivoting/bridging through a common language \cite{johnson2017google}. Given the lack of transfer across different \enxx translation tasks, it isn't hard to imagine why transferring across \yyent and \enxxt, to learn \yyxxt, is an even more challenging task.

Enabling direct translation between arbitrary languages has been widely studied, and has the potential to obviate the need for two-step pivoting which suffers from higher latency and accumulated errors. The most effective approach has been to simply synthesize parallel data \citep{firat2016zero,chen2017teacher,chen2018zero} and incorporate that into the training process. However, this two-stage approach becomes intractable when dealing with a large number of languages; the amount of synthesized data required to enable zero-shot translation grows quadratically with the number of languages. More recent work has demonstrated that direct translation quality may be improved even in a zero-shot setting, by incorporating more languages \citep{DBLP:journals/corr/abs-1903-00089}, adding regularization to promote cross-lingual transfer \cite{arivazhagan2019missing,gu2019improved,al2019consistency}, and modeling strategies that encourage shared cross-lingual representations \citep{lu2018neural,escolano2019towards}. 

We report the zero-shot performance of our $10$ language and $102$ language models, discussed in Section~\ref{sec:counter}, on selected language pairs in Table ~\ref{tab:zs}. We observe that zero-shot performance on similar languages, in this case \beru and \yide, is extremely high. We also notice that \textbf{the zero-shot performance for most language pairs increases as we move from the $10$ language model to the $102$ language model}, possibly due to the regularization effect in the capacity constrained setting, similar to what was observed in \cite{DBLP:journals/corr/abs-1903-00089}. This also indicates why methods that explicitly force languages to share the same representation space \cite{arivazhagan2019missing,gu2019improved,lu2018neural} may be key to improving zero-shot translation performance.

We next delve into pre-processing and vocabulary generation when dealing with hundreds of languages.

\section{Pre-processing and Vocabularies}
\label{sec:vocab}
Pre-processing and vocabulary construction lay central to the generalization ability of natural language processing systems.

To generalize to unseen examples, machine learning algorithms must decompose the input data into a set of basic units (e.g. pixels, characters, phonemes etc.) or building blocks. For text, this set of basic units or building blocks is referred to as the vocabulary. Given some text and a vocabulary, a segmentation algorithm or a pre-processor is applied to fragment the text to its building blocks upon which the learner may then apply inductive reasoning.

In essence, a properly defined vocabulary needs to (\textit{i}) maintain a high coverage by being able to compose most text and produce minimal number of Out-of-Vocabulary (OOV) tokens during segmentation, (\textit{ii}) have tractable size to limit computational and spatial costs, and (\textit{iii}) operate at the right level of granularity to enable inductive transfer with manageable sequence lengths which increase computational costs.

Early NMT models operated at the word level ~\citep{sutskever2014sequence,D14-1179,bahdanau2014neural}. Coverage issues arising from difficulty capturing all words of a language within a limited vocabulary budget promoted the development of character level systems~\citep{ling2015character,luong2016achieving,chung2016character,DBLP:journals/corr/Costa-JussaF16,lee2017fully,D18-1461}. These trivially achieve high coverage, albeit with the downside of increased computational and modeling challenges due to increased sequence lengths. Sub-word level vocabularies \citep{sennrich2016neural} have since found a middle ground and are used in most state-of-the-art NMT systems.

Constructing vocabularies that can model hundreds of languages with vast number of character sets, compositional units, and morphological variance is critical to developing massively multilingual systems, yet remains challenging. Early multilingual NMT models utilized \textit{separate} vocabularies for each language ~\cite{dong2015multi,luong2015multi,firat2016multi}; later ones used \textit{shared} multilingual vocabularies  ~\cite{sennrich2016neural,DBLP:journals/corr/HaNW16,johnson2017google} which are shared across languages. Recently, hybrid \textit{(Shared + Separate)} multilingual vocabularies and approaches for adding new languages to existing vocabularies~\cite{DBLP:journals/corr/abs-1811-01137} have also been explored. 

In this section we describe the simplistic approach to multilingual vocabulary construction used in our setup, inspect implicit characteristics of such vocabularies, and finally evaluate the downstream effects on multilingual machine translation.

\subsection{Vocabulary Construction and Characteristics}

We construct all our vocabularies using Sentence Piece Model (SPM)~\cite{kudo-richardson-2018-sentencepiece} to remove complexity that may arise from language specific pre-processing (e.g. tokenization, special character replacements etc.). The large number of languages used in our setup makes \textit{separate} per-language vocabularies infeasible. Therefore, we picked \textit{shared} sub-word vocabularies. While using a single shared vocabulary across all languages reduces complexity, it introduces other challenges that we discuss below.

With the large number of scripts introduced in a multilingual setting, the chance of a vocabulary producing unknowns (OOV) increases. Note that since only unrecognized characters will be encoded as OOV, if the vocabulary provides sufficient coverage over the alphabet from various scripts, OOV rates will be low. For SPMs, this is tuned using the character\_coverage option and we to a high value of $1-(5 * 10^{-6})$ which yields an alphabet size of around 12000, ensuring very low unknown rates for all the languages in our study.

\begin{figure}[t]
\centering
   \begin{subfigure}[t]{\textwidth}
   \includegraphics[width=0.48\textwidth]{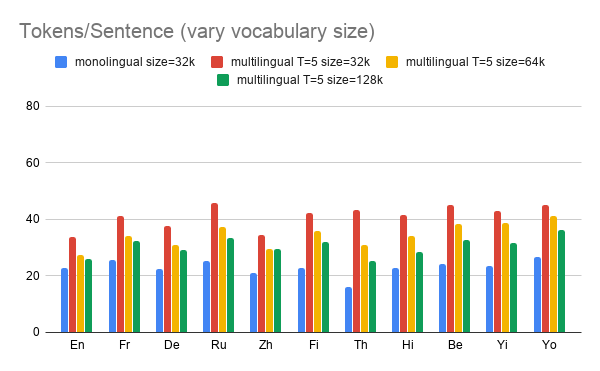}
\end{subfigure}
\begin{subfigure}[b]{\textwidth}
   \includegraphics[width=0.48\textwidth]{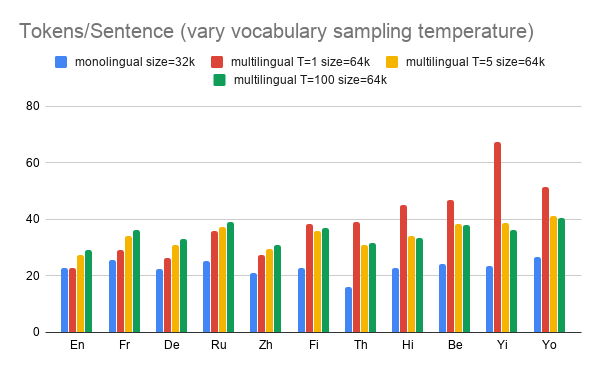}
\end{subfigure}
\caption{Average number of sentence-piece tokens per sentence on random samples drawn from the training set. We compare the increase in number of tokens per sentence for different languages, when moving from a standard monolingual vocabulary with 32k tokens to a multilingual vocabulary for which we vary the vocabulary size (size=$\{32k,64k,128k\}$ tokens) and the vocabulary sampling temperature ($T=\{1,5,100\}$).}
\label{fig:vocab_seqlen}
\end{figure}

While ensuring low unknown rates, the shift towards a vocabulary which largely consists of characters or short sub-words (in terms of the number of characters that they consist of) results in longer sequences after tokenization \cite{chung2016character,DBLP:journals/corr/Costa-JussaF16}. Longer sequence lengths, however, increase computational complexity and may also introduce optimization challenges due to longer range dependencies and require \textit{the learner} to model a more complex mapping function from a finer grained sequence to meaning. To avoid exceedingly long sequence lengths we need to increase the vocabulary size so that it may include longer tokens (sub-sequence that consist of longer number of characters).

Finally, a shared multilingual vocabulary runs the risk of favoring some languages over others, due to the imbalance of the dataset size the vocabulary is extracted. To reduce the effect of imbalanced dataset size we apply the same temperature sampling strategy discussed in Section~\ref{sec:learning} to Vocabulary Sampling.

In Figure~\ref{fig:vocab_seqlen} we report the effect of varying the vocabulary size (size=$\{32k,64k,128k\}$ tokens) and sampling temperature ($T=\{1,5,100\}$) on the average number of tokens per sentence on 10 indicative languages. We see that it is important to balance different languages by using a higher sampling temperature of $T=5$ or $T=100$, in order to avoid exceedingly long sequence lengths for low resource languages. Orthogonally, increasing the vocabulary size also helps to reduce sequence lengths, especially on low resource languages. Here we continue experiments with $T_V=5$ to stay aligned with our sampling approach during NMT training.

We next train and evaluate a few multilingual models to study the effect of different vocabulary sizes and sampling strategies on overall translation quality. Following the experiments in Section~\ref{sec:learning}, we train single multilingual models on all language pairs, using a data sampling temperature of $T=5$. All our models are Transformer-Big, trained with the same set of hyper-parameters, different only in terms of the vocabulary used. 

\begin{table}[t]
\centering
\begin{tabular}{l|c|c|c}
\enxx      & High 25  & Med. 52  & Low 25 \\ \hline
32k Vocab	& 27.69	& 16.84	& 12.90 \\\hline
64k Vocab	& 28.03	& 16.91	& 12.75\\\hline\hline
\xxen      & High 25  & Med. 52  & Low 25 \\ \hline
32k Vocab	& 33.24	& 29.40	& 26.18 \\\hline
64k Vocab	& 33.85	& 30.25	& 26.96 \\\hline
\end{tabular}
\caption{Average translation quality (BLEU) of multilingual models using different SPM vocabulary sizes, over different groups of languages. High 25 refers to the top 25 languages by dataset size, while low 25 refers to the bottom 25.}
\label{tab:vocabsize}
\end{table}

Table~\ref{tab:vocabsize} compares the quality of two models trained using vocabularies of size 32k and 64k. We notice that the model with the smaller 32k token vocab does noticeably worse on high resource languages when translating in both directions, and on \xxen translation in all resource settings. On the other hand, the smaller vocab model performs marginally better when translating into low resource languages on \enxx. For other medium resource languages, increased vocabulary size appears to be better on all directions. Our results here agree with existing literature \cite{D18-1461,DBLP:journals/corr/abs-1810-01480} suggesting that using smaller sub-word tokens, as is the case for smaller vocabularies, performs better in low resource settings due to improved generalization. Notable languages where the smaller vocabulary performs better include Corsican (\textit{co}) and Uzbek (\textit{uz}), both low resource languages which have known similar high resource languages to aid with generalization.

\begin{table}[t!]
\centering
\begin{tabular}{l|c|c|c}
\enxx      & High 25  & Med. 52  & Low 25 \\ \hline
$T_V=1$	& 27.81	& 16.72	& 12.73\\ \hline
$T_V=100$	& 27.83	& 16.86	& 12.78\\ \hline
$T_V=5$	& 28.03	& 16.91 &	12.75\\\hline\hline
\xxen      & High 25  & Med. 52  & Low 25 \\ \hline
$T_V=1$	& 33.82	& 29.78	& 26.27\\\hline
$T_V=100$	& 33.70	& 30.15	& 26.91\\\hline
$T_V=5$	& 33.85	& 30.25	& 26.96 \\\hline
\end{tabular}
\caption{Average translation quality (BLEU) of multilingual models using different sampling temperatures for vocabulary generation. High 25 refers to the top 25 languages by dataset size, while low 25 refers to the bottom 25.}
\label{tab:vocabsampling}
\end{table}

We compare the translation quality of models that vary only in their vocabulary sampling temperature in Table~\ref{tab:vocabsampling}. While not very pronounced, we do notice some differences in quality based on the vocabulary sampling temperature. Languages that perform better with a higher temperature of $T_V=5$ or $T_V=100$ include low-medium resource languages like Mongolian (\textit{mn}), Zulu (\textit{zu}), Corsican (\textit{co}) and Uzbek (\textit{uz}). These gains may have originated from two potential factors: (i) Smaller tokens for high resource languages result in better transfer to related low resource languages (ii) Better character coverage for languages with distinct character sets.

While the effects of vocabulary are much smaller than the trends observed for data sampling, failure to ensure careful character coverage and fair representation from all languages could nonetheless significantly impact translation quality.

So far in our study, we have presented our analysis and experimentation with data, training and vocabulary in multilingual scenarios. We next analyze the effect of several architectural choices on the quality of our massively multilingual NMT model.

\section{Modeling}
The quality of any neural network is largely dependent on its architecture and parametrization. The choice of the model, and the constraints on its parameters determine the extent of the transfer-interference trade-off, the learning dynamics and, ultimately, the performance limits of the model. In this section we analyze how choices regarding parameter sharing and model capacity can impact translation quality in the setting of massively multilingual NMT.

\subsection{Architecture}
In recent years, several revolutionary architectures have been developed to improve MT quality \cite{DBLP:journals/corr/GehringAGYD17,vaswani2017attention,chen-EtAl:2018:Long1,wu2019pay}. However, in the context of multilingual NMT, the most common prior imposed on models is typically in the form of (hard or soft) parameter sharing constraints across different languages pairs. In designing a multiingual NMT model, we would like to take advantage of the common structures and features shared by multiple languages, which imposes constraints on the model architecture. These constraints range from sharing a fixed length representation across all source and target pairs \cite{luong2015multi}, sharing small network modules, for example the attention parameters \cite{firat2016multi}, and sharing everything except the attention parameters \cite{ha2016toward,blackwood2018multilingual}, to sharing all parameters in the model across all language pairs \cite{johnson2017google,DBLP:journals/corr/HaNW16}. Some studies also explore partial parameter sharing strategies tailored for specific architectures \cite{sachan2018parameter,wang-EtAl:2018:EMNLP10}. More recently, with the increased number of languages \cite{Ye2018WordEmbeddings,DBLP:journals/corr/abs-1903-00089}, there has been work along the lines of applying soft sharing for MT, in the form of a shared set of meta-learned parameters used to generate task level parameters \cite{platanios2018contextual}. All of these sharing strategies come paired with their associated set of transfer-interference trade-offs, and regardless of the sharing strategy, 
in the absence of
a complete and accurate atlas of task relatedness, more transferrability also implies more interference. This phenomenon is known to be a common problem for multitask learning, and is also related to the stability vs plasticity dilemma \cite{carpenter1987art,Grossberg:1988:BBC:65669.104416}. See \cite{parisi18continual} for a categorization of such work.

Considering the scale of our experimental setup and end-goal, combinatorial approaches for parameter sharing do not emerge as plausible solutions. While ``learning to share'' approaches \cite{Kang:2011:LWS:3104482.3104548,ruder2017learning,platanios2018contextual} are promising alternatives, a more straightforward solution could be implicitly increasing the per-task capacity by increasing overall model capacity. Next we look into a brute-force approach to mitigate interference by enhancing model capacity.

\subsection{Capacity}
Over the last few years, scaling up model capacity has been shown to demonstrate huge improvements on several supervised and transfer learning benchmarks, including MT \cite{brock2018large,devlin2018bert,radford2018improving,shazeer2018mesh}. Scale often comes bundled with new hardware, infrastructure, and algorithmic approaches meant to optimize accelerator memory utilization and benefit faster computation, including methods like gradient checkpointing and low precision training \cite{courbariaux2014training,ott2018scaling}, memory efficient optimizers \cite{shazeer2018adafactor,gupta2018shampoo} and frameworks supporting model parallelism \cite{shazeer2018mesh,harlap2018pipedream,huang2018gpipe}.

While massive models have been shown to improve performance in single task settings \citep{shazeer2018mesh}, we would expect the gains to be even larger on a capacity-constrained massively multilingual task. We next try to quantify how the performance of our model scales with capacity. Additionally, we also compare two dimensions along which model capacity can be increased---depth and width---and compare how performance across different tasks is affected when scaling along these two dimensions.

We start with our Transformer-Big baseline with a 64k vocabulary, trained with a data sampling temperature of $T=5$. This model has around 400M parameters, including the embeddings and the softmax layers. We compare the performance of two scaled up models with around 1.3B parameters. Our first model is the wide model, with 12 layers in both the encoder and the decoder (24 layers in total), feed-forward hidden dimensions set to 16384, 32 attention heads and an attention hidden dimension set to 2048 \cite{shazeer2018mesh}. The deep model has 24 layers in the encoder and the decoder (48 layers in total), with all other hyper-parameters being equivalent to the Transformer-Big. To avoid trainability hurdles, both these models are trained with transparent attention \cite{bapna2018training}. Further, in order to enable training these massive models, we utilize GPipe \cite{huang2018gpipe} to incorporate efficient model parallelism. Both 1.3B param wide and deep models are trained with 128 TPUv3 chips, parallelized over 2 and 4 cores respectively.\footnote{Note, each TPUv3 chip has 2 cores.} We use the same 4M token batch size used for all our multilingual experiments.

\begin{figure}[t!]
\centering
   \begin{subfigure}[t]{\textwidth}
   \includegraphics[width=0.48\textwidth]{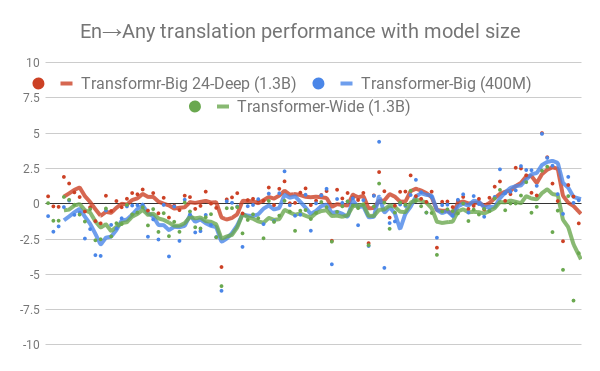}
\end{subfigure}
\begin{subfigure}[b]{\textwidth}
   \includegraphics[width=0.48\textwidth]{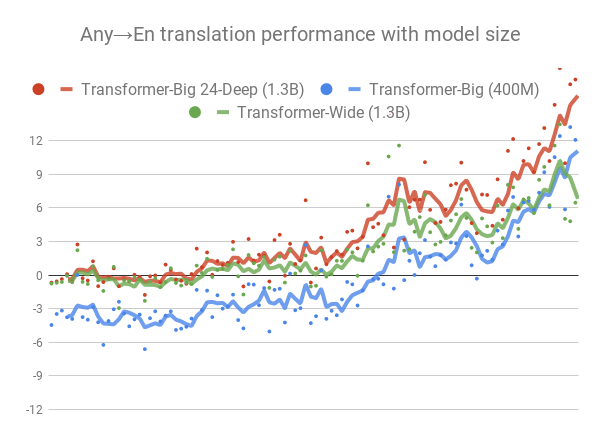}
\end{subfigure}
\caption{Effect of increasing capacity on the performance of multilingual models. From left to right, languages are arranged in decreasing order of available training data. Results are reported relative to those of the bilingual baselines (\ref{fig:bilingual}). The plots correspond to the following models:  blue: 400M param `Transformer-Big', green: 1.3B param, 12 layer wide model and red: 1.3B param, 24 layer model. Best viewed in color.}
\label{fig:capacity}
\end{figure}

\begin{table}[t!]
\centering
\begin{tabular}{l|c|c|c}
\enxx      & High 25  & Med. 52  & Low 25 \\ \hline
Bilingual	& 29.34	& 17.50	& 11.72 \\\hline 
400M&	28.03&	16.91 &	12.75 \\\hline
1.3B Wide&	28.36&	16.66 &	11.14\\\hline
1.3B Deep&	29.46&	17.67 &	12.52 \\\hline\hline
\xxen      & High 25  & Med. 52  & Low 25 \\ \hline
Bilingual	& 37.61	& 31.41	& 21.63 \\\hline
400M&	33.85 &	30.25 &	26.96 \\\hline
1.3B Wide&	37.13&	33.21&	27.75 \\\hline
1.3B Deep&	37.47&	34.63&	31.21 \\\hline
\end{tabular}
\caption{Average translation quality (BLEU) of multilingual models with increasing capacity. High 25 refers to the top 25 languages by dataset size, while low 25 refers to the bottom 25.}
\label{tab:capacity}
\end{table}

The performance of these two models and the baseline Transformer-Big is depicted in Figure~\ref{fig:capacity} (also summarized in Table~\ref{tab:capacity}). We notice that both these models improve performance by significant amounts on the high resource languages, when compared against the capacity constrained massively multilingual baseline (blue curves in Figure~\ref{fig:capacity}). However, the deep model handily beats both, the baseline and the equivalent capacity wide model, by significant margins on most of the language pairs. We also notice that, unlike the wide model, the deep model does not overfit in low resource languages and, in fact, significantly enhances transfer to low resource languages on the \xxen translation tasks.

Our results suggest that model capacity might be one of the most important factors determining the extent of the transfer-interference trade-off. However, naively scaling capacity might result in poor transfer performance to low resource languages. For example, our wide Transformer while significantly improving performance on the high resource languages, fails to show similar gains in the low resource setting. While deeper models show great performance improvements, they come bundled with high decoding latency, a significantly larger computational footprint, and trainability concerns including vanishing/exploding gradients, early divergence, ill-conditioned initial conditions etc. \cite{scalingispredictable}. Further research into various aspects of scalability, trainability and optimization dynamics is expected to be a fruitful avenue towards universal NMT.

We next delve into evaluation challenges posed by multilingual models.

\section{Evaluation}
Metrics for automatic quality evaluation \cite{papineni-EtAl:2002:ACL} have been critical to the rapid progress in machine translation, by making evaluation fast, cheap and reproducible. For multilingual NMT, new concerns arise due to the multi-objective nature of the problem and the inherent quality trade-offs between languages.

Inter-language quality trade-offs arise due to various decisions made while constructing, training and selecting a model. When constructing the model, the vocabulary may be constructed to favor a certain script or group of languages, or the language specific parameters may be unevenly distributed across languages. During training, the optimization settings or the rate at which training data from different languages are sampled strongly influence the eventual performance on different languages. Finally, when selecting a checkpoint\footnote{Certain snapshot of the model parameters.}, the model may perform better on high resource languages at later checkpoints but may have regressed on low resource languages by that time due to over-training (or under-training for the opposing case). Each of these choices naturally may favor certain languages over others. 

To choose between the aforementioned trade-offs, we need a translation quality metric that is both effective and comparable across languages. This in and of itself is a hard problem with an ever growing list of hundreds of metrics to choose from. Oftentimes these metrics vary in their effectiveness across languages; WMT shared tasks \cite{ma2018results} report that the specific language, dataset, and system significantly affect which metric has the strongest correlation with human ratings. When metrics are sufficiently effective across languages they are not always comparable. N-gram based metrics \cite{papineni-EtAl:2002:ACL,doddington2002automatic,wang2016character,popovic2015chrf} that measure lexical overlap require tokenization which is highly affected by language specific factors such as alphabet size and morphological complexity. In fact, even within the same language, tokenization discrepancies pose significant challenges to reliable evaluation \cite{post2018call}. Embedding based approaches \cite{stanojevic-simaan-2014-beer} may be language agnostic and help address these issues.

Equally important to choosing a metric is choosing an evaluation set. Most existing metrics are not consistent \cite{banerjee2005meteor} and for the same model vary based on the domain or even the specific sentences that they are being evaluated on. For example, there are significant differences of 3-5 BLEU between the WMT dev and test sets for the same language pair. Such consistency issues may further exacerbate the difficulty of comparing system quality across languages if we use different test sets for each language. This may be addressed by ensuring that evaluation is performed on the same corpora that has been translated to all the languages, i.e. multi-way parallel data. Even so, attention must be paid to the original language \cite{freitag2019text} and domain such data is collected from.

\section{Open Problems in Massively Multilingual NMT}
\paragraph{Data and Supervision} 
Whereas we focus solely on supervised learning, for many low resource languages it becomes essential to learn from monolingual data. There has been a lot of recent work on incorporating monolingual data to improve translation performance in low and zero resource settings, including research on back-translation \cite{sennrich2015improving,edunov2018understanding}, language model fusion \cite{gulcehre2015using,sriram2017cold}, self-supervised pre-training \cite{dai2015semi,ramachandran2016unsupervised,zhang2016exploiting,song2019mass} and unsupervised NMT \cite{lample2017unsupervised,artetxe2017unsupervised}. Languages where large swathes of monolingual data are not easily available might require more sample efficient approaches for language modeling, including grounding with visual modalities or sources of information \cite{huang2016attention}, and learning from meta-data or other forms of context. Being able to represent and ground information from multiple modalities in the same representational space is the next frontier for ML research.

The scope of our study is limited to $103$ languages, a minuscule fraction of the thousands of existing languages. The heuristics and approaches we develop will be less and less applicable as we include more languages and incorporate other forms of data. As we scale up model sizes and the number of languages learned within a single model, approaches that require multi-stage training or inference steps will likely become infeasible. Work towards better integration of self-supervised learning with the supervised training process is more likely to scale well with larger model capacities and increasing number of languages.

\paragraph{Learning} Developing learning approaches that work well for multitask models is essential to improving the quality of multilingual models. Our analysis from Section~\ref{sec:learning} demonstrates that even simple heuristics for data sampling/balancing can significantly improve the extent of transfer and interference observed by individual tasks. However, our heuristic strategy only takes dataset size into account when determining the fraction of per-task samples seen by the model. Research on exploiting task-relatedness \cite{lee2016asymmetric,neubig2018rapid}, curriculum learning on noisy data \cite{van2017dynamic,wang2018denoising}, and automated curriculum learning from model state \cite{bengio2009curriculum,graves2017automated} have demonstrated success in multitask learning, including for NMT. Other relevant threads of work include research on meta-learning to learn model hyper-parameters \cite{nichol2018first,baydin2017online}, model parameters \cite{ha2016hypernetworks,platanios2018contextual} and models that learn new tasks with high sample efficiency \cite{finn2017model} without forgetting existing tasks or languages \cite{rusu2016progressive,kirkpatrick2017overcoming,DBLP:journals/corr/abs-1811-01137}. When scaling to thousands of languages, approaches that can automatically learn data sampling, curricula, model hyper-parameters and parameters, to train models that quickly adapt to new tasks, is again expected to become increasingly important. 

\paragraph{Increasing Capacity} Increasing the model capacity has been demonstrated to be a sure-shot approach to improving model quality in the presence of supervised data for several tasks, including Image Generation \cite{brock2018large}, Language Modeling and transfer learning \cite{radford2018improving,devlin2018bert} and NMT \cite{shazeer2018mesh}. Validating this trend, one key result from our study is the need for sufficient model capacity when training large multitask networks. Other than the systems and engineering challenges \cite{46078,huang2018gpipe,shazeer2018mesh}, training deep and high capacity neural networks poses significant trainability challenges \cite{montavon2018methods}. A better theoretical understanding of the generalization ability of deep and wide models \cite{raghu2017expressive,arora2018optimization}, trainability challenges including exploding and vanishing gradients \cite{glorot2010understanding,balduzzi2017shattered} and empirical approaches to counter these challenges \cite{bapna2018training,zhang2019fixup} are critical to further scaling of model capacity.

\paragraph{Architecture and Vocabulary} Extending existing approaches for vocabulary construction and neural modeling to work better in multitask settings \cite{Ma:2018:MTR:3219819.3220007,houlsby2019parameter} in order to strike the right balance between shared and task-specific capacity, or learning network structure in order to maximally exploit task relatedness \cite{li2019learn} are exciting avenues for future research. As we scale up to thousands of languages, vocabulary handling becomes a significantly harder challenge. Successfully representing thousands of languages might require character \cite{lee2017fully,D18-1461}, byte \cite{gillick2015multilingual} or even bit level modeling. Modeling extremely long sequences of bit/byte-level tokens would present its own set of modeling challenges. Finally, while scaling up existing approaches is one way to improve model quality, some approaches or architectures are more efficient to train \cite{shazeer2017outrageously,vaswani2017attention,wu2019pay}, more sample efficient, or faster at inference \cite{gu2017non,roy2018theory}. As models get larger, improving training and inference efficiency become more important to keep training and inference times within reasonable limits from both a practical and environmental perspective  \cite{strubell2019energy}. 

\section{Conclusion}

Although we believe that we have achieved a milestone with the present study, building on five years of multilingual NMT research, we still have a long way to go towards truly universal machine translation. 
In the open problems and future directions enumerated above, many promising solutions appear to be interdisciplinary, making multilingual NMT a plausible general test bed for other machine learning practitioners and theoreticians.

\section*{Acknowledgments}
We would like to thank the Google Translate and Google Brain teams for their useful input and discussion, and the entire Lingvo development team for their foundational contributions to this project. We would also like to thank Katherine Lee, Thang Luong, Colin Raffel, Noam Shazeer, and Geoffrey Hinton for their insightful comments. \\

\bibliography{tacl2018v2}
\bibliographystyle{acl_natbib}

\appendix

\begin{table*}[]
\centering
\begin{tabular}{ll|ll|ll}

\hline
Language         & Id  & Language      & Id  & Language      & Id \\ \hline \hline
Afrikaans        & af  & Hebrew        & iw  & Polish        & pl \\
Albanian         & sq  & Hindi         & hi  & Portuguese    & pt \\
Amharic          & am  & Hmong         & hmn & Punjabi       & pa \\
Arabic           & ar  & Hungarian     & hu  & Romanian      & ro \\
Armenian         & hy  & Icelandic     & is  & Russian       & ru \\
Azerbaijani      & az  & Igbo          & ig  & Samoan        & sm \\
Basque           & eu  & Indonesian    & id  & Scots\_Gaelic & gd \\
Belarusian       & be  & Irish         & ga  & Serbian       & sr \\
Bengali          & bn  & Italian       & it  & Sesotho       & st \\
Bosnian          & bs  & Japanese      & ja  & Shona         & sn \\
Bulgarian        & bg  & Javanese      & jw  & Sindhi        & sd \\
Burmese          & my  & Kannada       & kn  & Sinhalese     & si \\
Catalan          & ca  & Kazakh        & kk  & Slovak        & sk \\
Cebuano          & ceb & Khmer         & km  & Slovenian     & sl \\
Chinese          & zh  & Korean        & ko  & Somali        & so \\
Corsican         & co  & Kurdish       & ku  & Spanish       & es \\
Croatian         & hr  & Kyrgyz        & ky  & Sundanese     & su \\
Czech            & cs  & Lao      & lo  & Swahili       & sw \\
Danish           & da  & Latvian       & lv  & Swedish       & sv \\
Dutch            & nl  & Lithuanian    & lt  & Tajik         & tg \\
Esperanto        & eo  & Luxembourgish & lb  & Tamil         & ta \\
Estonian         & et  & Macedonian    & mk  & Telugu        & te \\
Filipino/Tagalog & tl  & Malagasy      & mg  & Thai          & th \\
Finnish          & fi  & Malay         & ms  & Turkish       & tr \\
French           & fr  & Malayalam     & ml  & Ukrainian     & uk \\
Frisian          & fy  & Maltese       & mt  & Urdu          & ur \\
Galician         & gl  & Maori         & mi  & Uzbek         & uz \\
Georgian         & ka  & Marathi       & mr  & Vietnamese    & vi \\
German           & de  & Mongolian     & mn  & Welsh         & cy \\
Greek            & el  & Nepali        & ne  & Xhosa         & xh \\
Gujarati         & gu  & Norwegian     & no  & Yiddish       & yi \\
Haitian\_Creole  & ht  & Nyanja        & ny  & Yoruba        & yo \\
Hausa            & ha  & Pashto        & ps  & Zulu          & zu \\
Hawaiian         & haw & Persian       & fa  &               &    \\ \hline
\end{tabular}
\caption{List of BCP-47 language codes used throughout this paper \cite{bcp47}.}.
\label{tab:lang_ids}
\end{table*}

\end{document}